\documentclass[letterpaper]{article} 
\usepackage{aaai2027}  
\usepackage[hyphens]{url}  
\usepackage{graphicx} 
\usepackage{natbib}  
\usepackage{caption} 
\usepackage{algorithm}
\usepackage{algorithmic}
\usepackage[switch]{lineno}
\usepackage{amsfonts}
\usepackage{amsmath}
\usepackage{booktabs}
\usepackage{multirow}
\usepackage{booktabs}
\usepackage{subcaption}

\usepackage{newfloat}
\usepackage{listings}
\DeclareCaptionStyle{ruled}{labelfont=normalfont,labelsep=colon,strut=off} 
\floatstyle{ruled}
\newfloat{listing}{tb}{lst}{}
\floatname{listing}{Listing}

\usepackage{booktabs}

\title{Revisiting Cross-Reconstruction for Generalizable Deepfake Detection}
\author{
    Bingjian Yang\textsuperscript{\rm 1},
    Shilei Zhao\textsuperscript{\rm 1},
    Zheng Wang\textsuperscript{\rm 1}\corresponding
}
\affiliations{
    \textsuperscript{\rm 1}National Engineering Research Center for Multimedia Software, School of Computer Science, Wuhan University, China\\
    yangbingjian@whu.edu.cn, 649162208@qq.com, wangzwhu@whu.edu.cn
}

\begin{document}

\maketitle

\begin{abstract}

Existing image forgery detectors often suffer from generalization to unseen manipulation methods due to the limited ability to capture transferable forensic cues. 
Recent cross-reconstruction based methods attempt to improve generalization through semantic-artifact disentanglement, but typically align heterogeneous artifacts across generators and exclude artifact representations during reconstruction, which may overlook the inherent diversity and visual cues of manipulation artifacts.
In this work, we revisit cross-reconstruction and introduce an artifact-oriented disentanglement framework for robust image forgery detection. 
We argue that \textbf{artifact diversity}, i.e., the intrinsic variations of manipulation artifacts introduced by different generation processes, contains complementary forensic cues rather than undesirable domain variations. 
Instead of enforcing explicit artifact alignment, our framework preserves diverse artifact characteristics through semantically aligned cross-generator reconstruction. 
Furthermore, we incorporate artifact representations into the reconstruction process and introduce a masked frequency-aware reconstruction strategy to emphasize manipulation-related residuals while reducing semantic interference. 
This design enables the model to learn transferable forensic representations from diverse artifacts.
Extensive experiments on multiple benchmark datasets demonstrate improvements under both cross-dataset and cross-generator evaluation settings. 
Further analysis and ablation studies validate the effectiveness of artifact diversity preservation and artifact-aware cross-reconstruction.

\end{abstract}

\section{Introduction}

\begin{figure}[t]
    \centering

    \begin{minipage}{0.47\linewidth}
        \centering
        \includegraphics[width=0.85\linewidth]{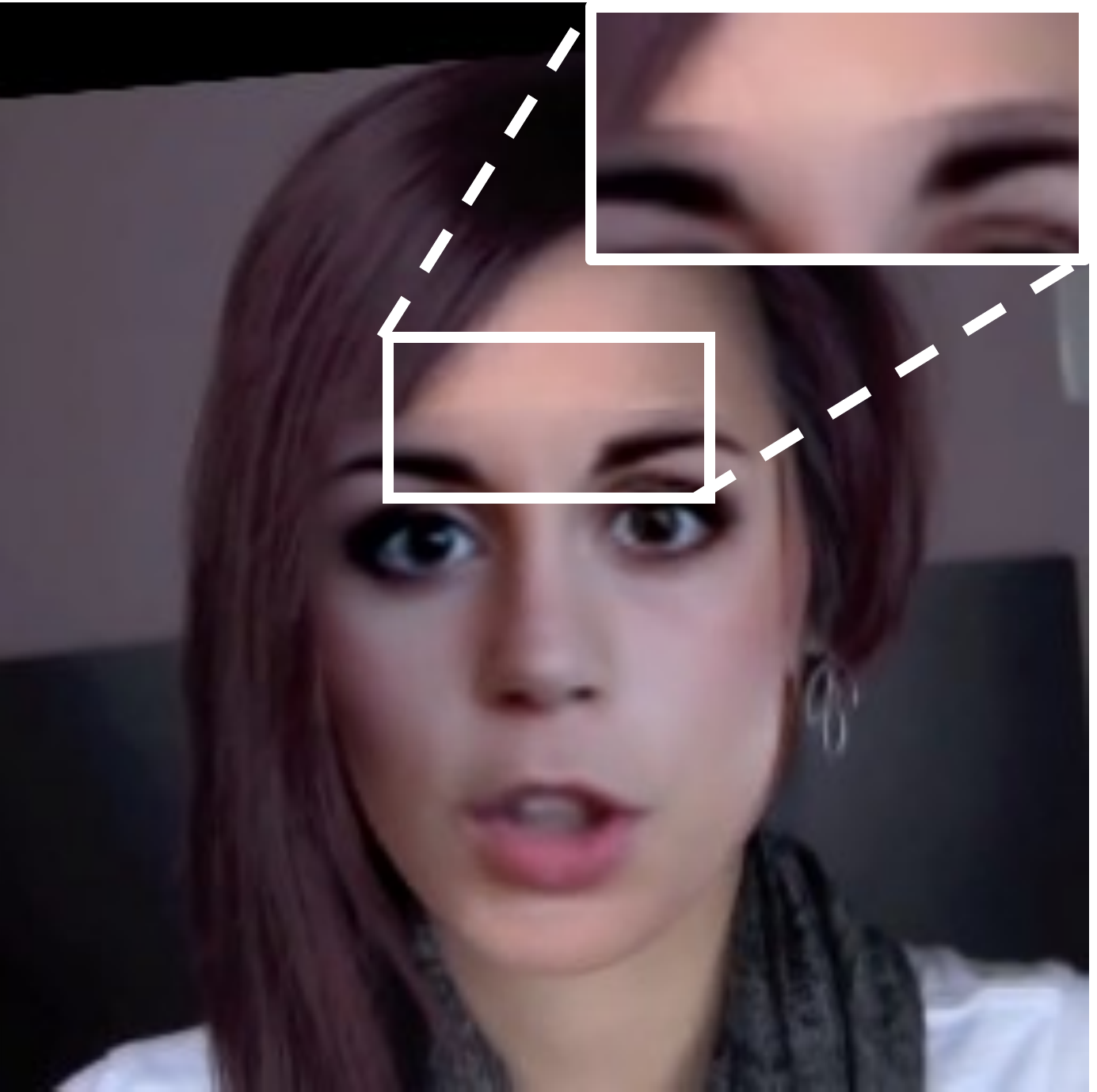}
        
        \small (a)
    \end{minipage}
    \hfill
    \begin{minipage}{0.47\linewidth}
        \centering
        \includegraphics[width=0.85\linewidth]{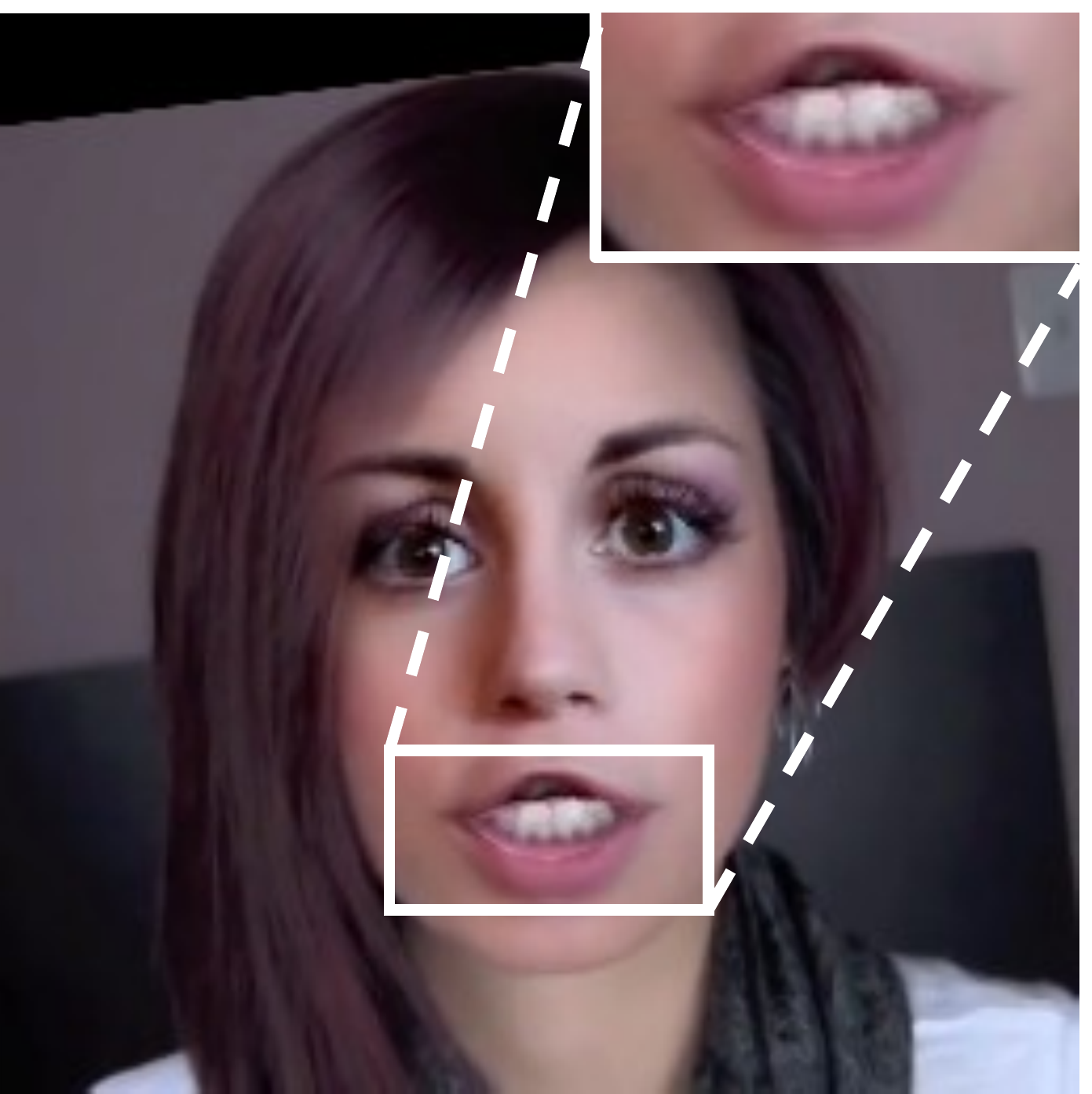}
        
        \small (b)
    \end{minipage}
    
    \begin{minipage}{0.95\linewidth}
        \centering
        \includegraphics[width=0.75\linewidth]{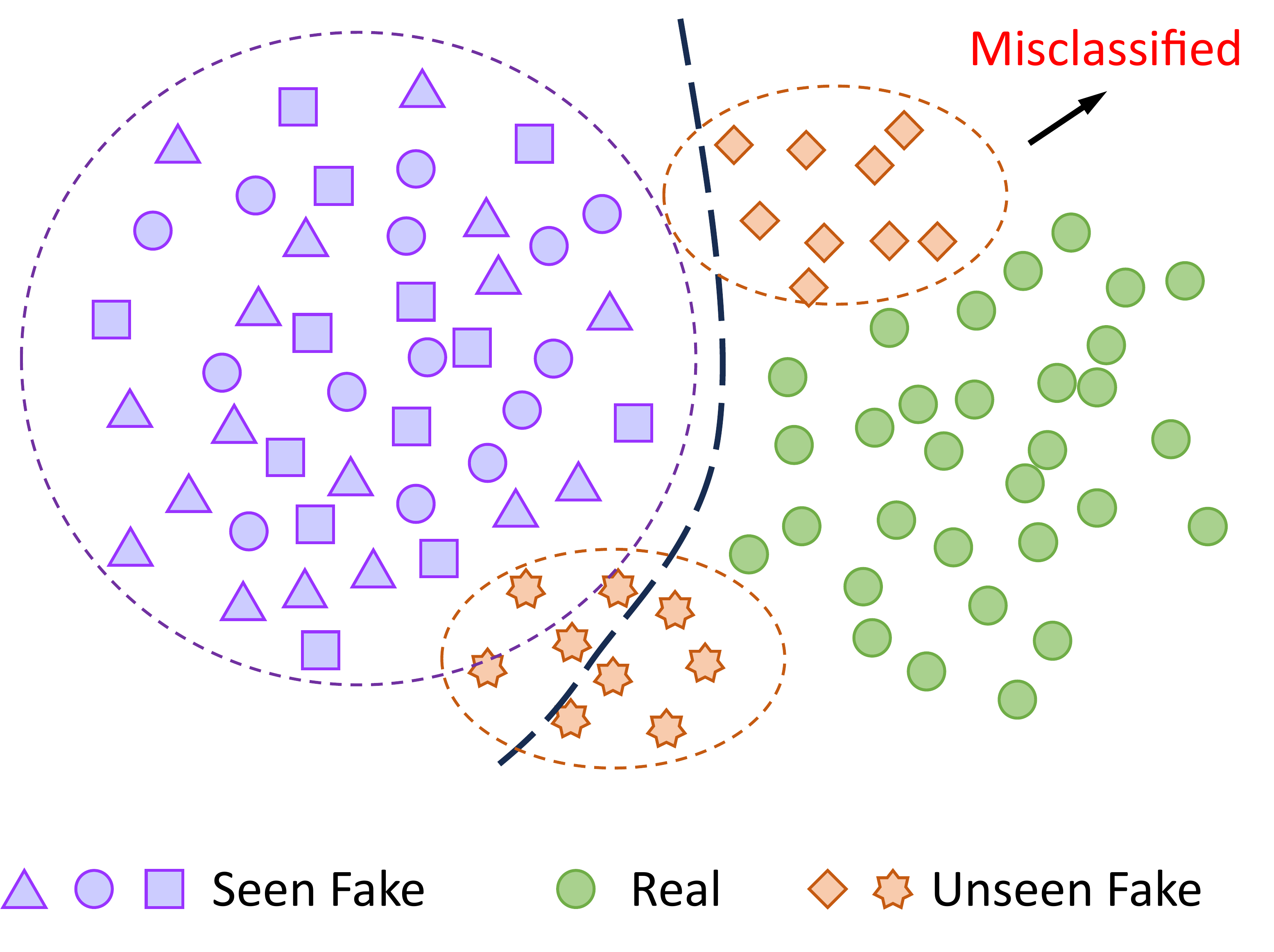}
        
        \small (c)
    \end{minipage}

    \vspace{0.2em}

    \caption{Motivation of preserving artifact diversity. 
(a)-(b) Examples from different generators, showing diverse forgery artifacts, including color inconsistency and unnatural facial details (e.g., teeth). 
(c) Artifact alignment across generators may harm generalization to unseen generators, causing misclassification of unseen manipulations.}
    \label{fig:motivation}
\end{figure}
Recent image forgery detectors have increasingly explored disentangled representation learning to separate semantic information from manipulation artifacts for better generalization. 
Among them, cross-reconstruction based methods \cite{yan2023ucf,lin2024preserving} construct paired real-fake samples and decompose their representations into semantic and artifact components. By swapping artifact representations while preserving semantic representations, they reconstruct the original RGB images to encourage semantic-artifact disentanglement. They further divide artifact representations into generator-related and generator-invariant components, where the former is supervised by generator classification and the latter is utilized for forgery detection.

However, we argue that such paradigm overlooks an important property of forgery artifacts, namely \textbf{artifact diversity} as shown in Fig.~\ref{fig:motivation}.  
Artifact diversity refers to the intrinsic variations of manipulation artifacts introduced by different generation models, training strategies, and synthesis pipelines. 
Although these artifacts exhibit heterogeneous distributions across generators, they often contain complementary forensic cues that reveal different aspects of the manipulation process. 
Therefore, artifact diversity should not be regarded as undesirable domain variation that needs to be eliminated, but rather as valuable forensic knowledge that should be preserved and exploited.

As illustrated in Fig.~\ref{fig:motivation}, existing cross-reconstruction methods tend to align artifacts from different generators into a shared representation space. 
While this strategy improves invariance to seen generators, it may suppress generator-specific but transferable forensic cues and lead to overfitting. 
Consequently, the learned representation can be biased toward the artifact statistics of training generators and fail to capture the diverse forensic patterns required for unseen manipulations.
Besides artifact alignment, another limitation lies in the role of artifacts during reconstruction. 
Existing methods typically exclude artifact representations from reconstruction process, assuming that artifacts are merely reconstruction-irrelevant noise. 
However, forgery artifacts are not merely reconstruction-irrelevant noise. As illustrated in Fig.~\ref{fig:motivation}(a)-(b), they contain meaningful visual information, such as color mismatch and detail distortion, providing complementary forensic cues for identifying manipulated images. 
Therefore, whether artifacts should participate in reconstruction remains an open question.

To address these limitations, we propose an artifact-oriented disentanglement framework for robust image forgery detection. 
Instead of enforcing explicit artifact alignment, our method preserves artifact diversity and learns transferable forensic representations from heterogeneous artifacts. 
We redesign cross-reconstruction by incorporating artifact representations as reconstruction guidance. 
To enhance artifact supervision, we introduce a masked frequency-aware reconstruction strategy that suppresses semantic-dominant components while emphasizing manipulation-related residuals.
Extensive experiments on multiple benchmarks demonstrate consistent improvements under cross-dataset settings. 
Ablation studies further validate the effectiveness of artifact diversity preservation and artifact-aware reconstruction.
Our contributions are summarized as follows:
\begin{itemize}
    \item We revisit cross-reconstruction for deepfake detection and identify two limitations: aligning heterogeneous artifacts suppresses artifact diversity, while the exclusion of artifact representations from reconstruction overlooks some forensic cues.

    \item We propose an artifact-aware disentanglement framework using semantically aligned cross-generator pairs. Cross-reconstruction incorporates artifact representations into the reconstruction process, enabling the exploitation of diverse manipulation cues. Meanwhile, masked high-frequency residual reconstruction emphasizes manipulation-related details and promotes transferable forensic representation learning.
    
    \item Extensive cross-dataset experiments demonstrate competitive state-of-the-art performance. Ablation studies further validate the benefits of artifact diversity preservation and artifact-aware cross-reconstruction.
\end{itemize}

\section{Related Works}

\subsection{Generalizable Deepfake Detection}
Existing deepfake detectors \cite{li2023spatio,ba2024exposing,concas2024quality} have achieved remarkable performance under in-domain evaluation, yet their generalization ability remains limited when encountering unseen manipulation methods or datasets. To alleviate this issue, recent studies\cite{zhu2026unleashing,chou2026improving} have explored various strategies for learning more generalizable representations. Some works ~\cite{stehouwer2019detection,li2020face,nguyen2024laa} exploit forgery artifacts in the spatial and frequency domains, such as blending inconsistencies, abnormal noise patterns, and spectral artifacts. Another direction introduces large-scale pre-trained vision models, such as Xception\cite{chollet2017xception} and CLIP\cite{radford2021learning} into deepfake detection to leverage their powerful visual representations and superior transferability. Effort \cite{yan2024effort} explicitly enforces orthogonality between semantic and forgery subspaces based on CLIP representations. CFB \cite{li2025critical} performs multi-scale feature-level disentanglement of semantic and artifact representations using an Xception backbone. Beyond semantic-artifact disentanglement, recent studies\cite{lin2024preserving,yang2026improving} further decompose forensic representations into generator-specific and generator-shared components to improve cross-generator generalization. Such methods assume that the shared artifact space can be effectively learned from a limited number of seen generators. However, given the limited diversity of generators available during training and the rapid evolution of deepfake generation techniques, this assumption may be overly restrictive and potentially lead to over-specialized representations. We therefore argue that preserving the diversity of forgery artifacts, rather than enforcing a predefined shared artifact space, may provide a more effective path toward robust generalization.

\subsection{Reconstruction Based Representation Learning in Deepfake Detection}
Reconstruction based representation learning\cite{kim2018disentangling,lee2018diverse, PhyTrace} has been widely adopted to disentangle latent factors and learn discriminative feature representations.
By reconstructing samples from latent representations, these methods encourage different representation subspaces to capture complementary information while suppressing nuisance factors.
Inspired by its success in representation disentanglement, recent deepfake detection methods have incorporated reconstruction to separate semantic content from forgery-related representations.
UCF \cite{yan2023ucf} introduces cross-reconstruction to uncover common forgery features shared across different manipulation methods, encouraging the detector to focus on manipulation-relevant information. 
Similarly, PFG\cite{lin2024preserving} employs cross-reconstruction to disentangle demographic attributes from forgery representations, thereby learning domain-agnostic forensic features for improved cross-domain generalization.
 CFB\cite{li2025critical} extends reconstruction from the image level to the feature level, enabling multi-scale disentanglement of semantic and artifact representations.
Despite their effectiveness, the reconstruction objective is designed in a task-agnostic manner rather than being tailored for forensic representation learning. Given that forgery artifacts are predominantly manifested in high-frequency components, directly reconstructing RGB images may provide suboptimal supervision for learning artifact-aware representations.

\section{Method}

\subsection{Overview}



Our framework aims to learn transferable forensic representations by preserving the diversity of manipulation artifacts rather than enforcing explicit artifact alignment. Given semantically aligned image pairs generated by different manipulation methods, the framework extracts disentangled semantic and artifact representations through a query-based encoder and a decouple head. To exploit the complementary information in diverse artifacts, we introduce a diversity-preserving cross-reconstruction module, where artifact representations are exchanged and incorporated into reconstruction. Meanwhile, masked high-frequency residuals are adopted as reconstruction targets to emphasize manipulation-related variations and facilitate the learning of transferable forensic representations.

For forgery detection, we introduce a CLIP-based classification module guided by artifact-aware attention biases. The entire framework is jointly optimized with classification loss, artifact contrastive loss, and cross-reconstruction loss, enabling effective exploitation of diverse manipulation cues for generalized deepfake detection.
\subsection{Training Pair Formation}
\label{sec:pair_formation}
\begin{figure}[!ht]
    \centering
    \includegraphics[width=\linewidth]{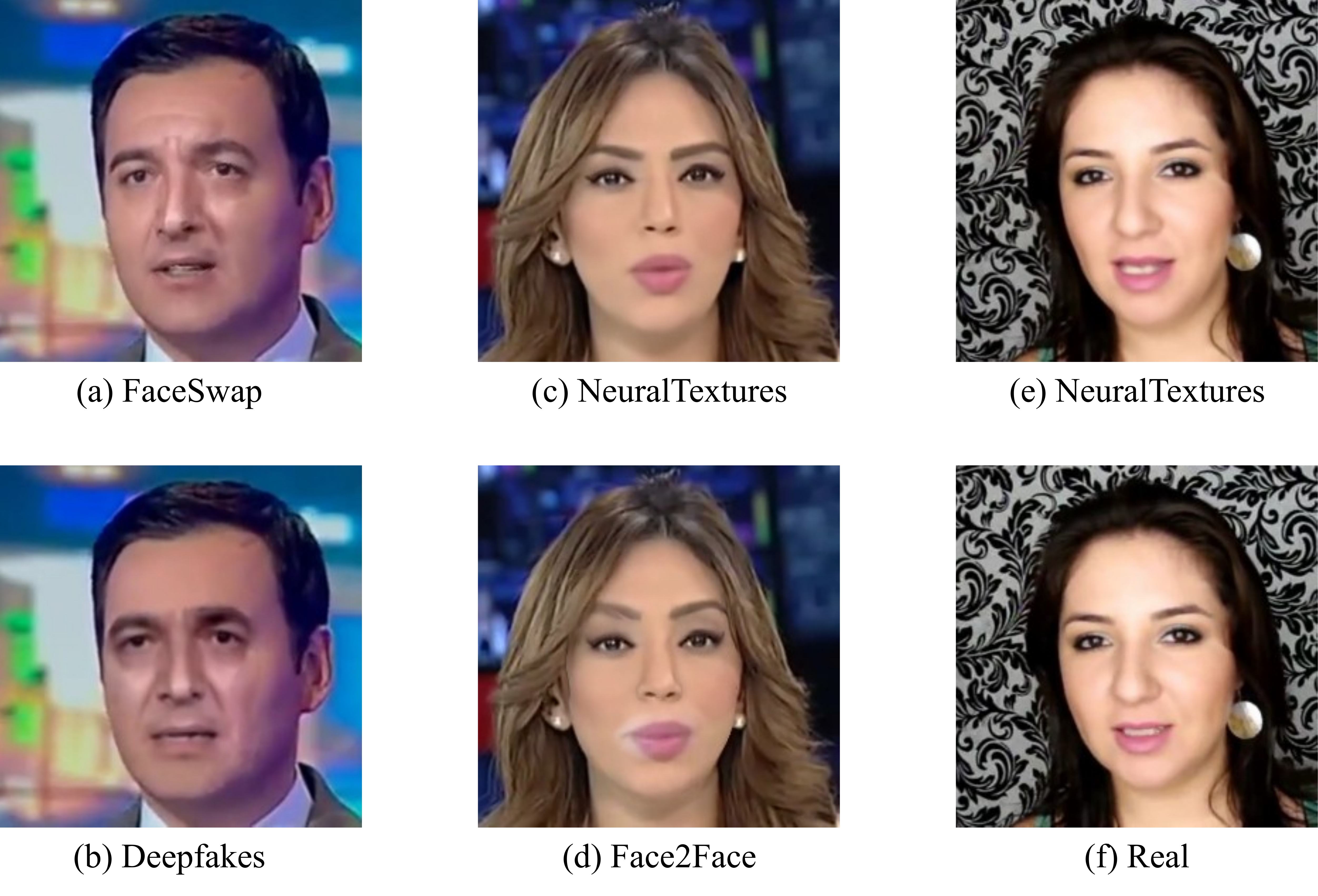}
    \caption{Examples of the training data. (a) and (b) are generated by two face swapping methods. (c) and (d) are generated by two face reenactment methods. (e) and (f) show a face reenactment sample and its corresponding pristine image, which preserves the same semantic content.}
    \label{fig:pair_construction}
\end{figure}

We adopt the FaceForensics++ \cite{rossler2019faceforensics++} dataset as the training set, since it provides pristine videos together with multiple face manipulation methods generated from the same source videos. Different from conventional cross-reconstruction methods, which typically construct training pairs from real and fake images, our training pairs are composed of semantically similar samples generated by different manipulation methods, as illustrated in Fig.~\ref{fig:pair_construction}. Specifically, FF++ dataset consists of pristine videos together with two categories of manipulated videos: face reenactment and face swapping.
During pair construction, pristine videos and face reenactment videos are grouped together, while face swapping videos are treated as a separate group. For each training pair, two videos are sampled from one group. 
The paired samples therefore exhibit highly consistent facial pose, expression, and background while originating from different generators. Afterward, frames with identical indices are extracted from each paired video to form image pairs for training. Standard face preprocessing, including face detection, alignment, and cropping, follows the protocol in DeepfakeBench \cite{yan2023deepfakebench}. 

\begin{figure*}[t]
    \centering
    \includegraphics[width=\linewidth]{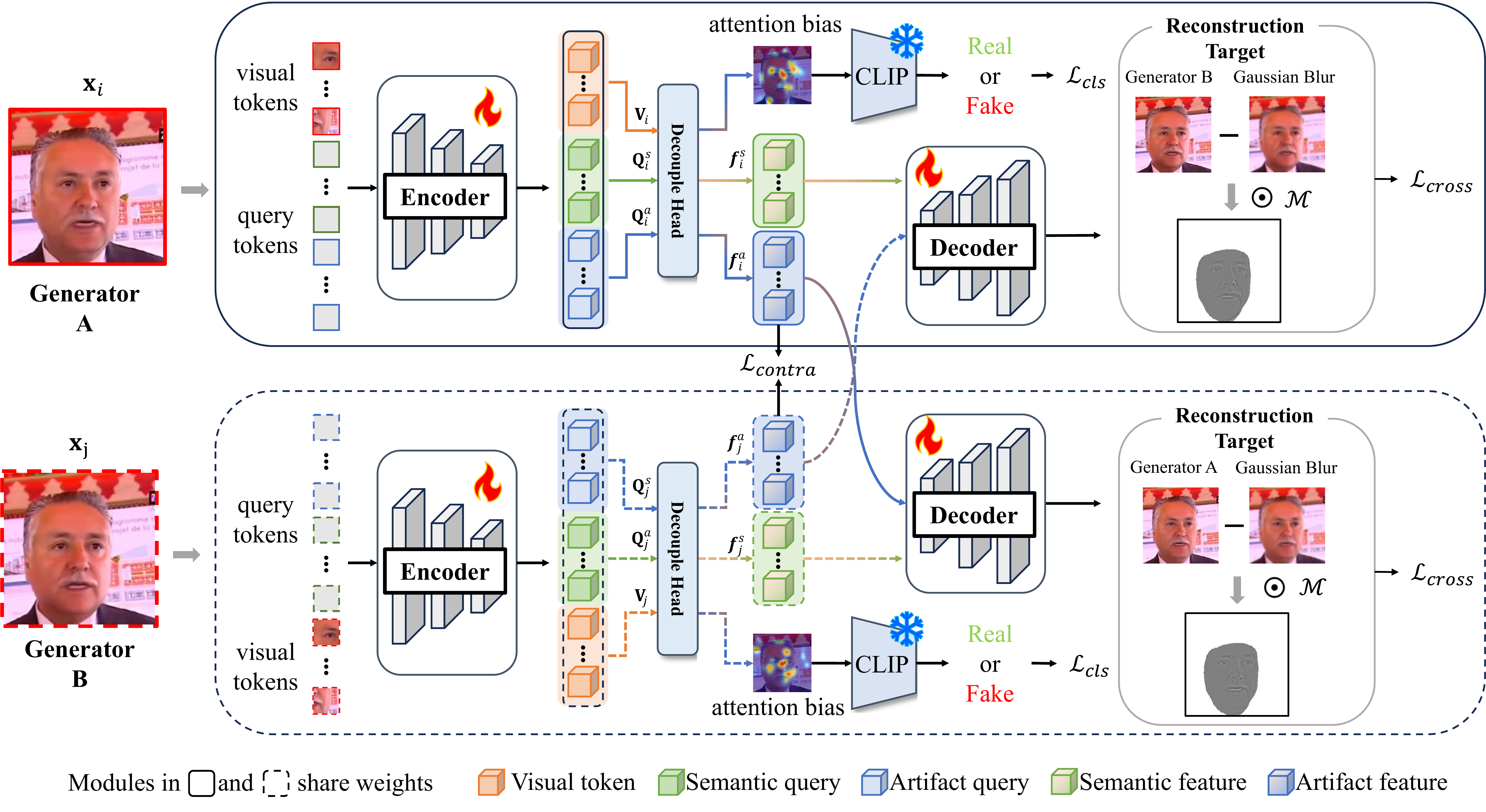}
    \caption{Overview of the proposed artifact-aware disentanglement framework. Given a pair $(x_i,x_j)$, the encoder extracts visual tokens $\mathbf{V}$ and learns query tokens $\mathbf{Q}^s$ and $\mathbf{Q}^a$ . The decouple head produces semantic features $\boldsymbol{f}^{s}$ and artifact features $\boldsymbol{f}^{a}$, which are used for cross-reconstruction. The artifact-aware attention bias guides the frozen CLIP encoder for forgery prediction. In the decoder, artifact features are exchanged between paired samples through AdaIN-based fusion to reconstruct the high-frequency masked targets. $\mathcal{M}$ denotes the union of manipulation masks, and $\mathcal{L}_{cls}$, $\mathcal{L}_{contra}$, and $\mathcal{L}_{cross}$ represent classification loss, artifact contrastive loss, and cross-reconstruction loss, respectively.}
    \label{fig:pipeline}
\end{figure*}

\subsection{Diversity Preserving Cross-Reconstruction}
\subsubsection{Encoder.}

Our encoder is built upon a lightweight Vision Transformer (ViT). Following ForensicAdapter~\cite{cui2025forensics} and VLA-Forge~\cite{zhu2026unleashing}, we introduce a set of learnable query tokens as $\mathbf{Q_0}\in\mathbb{R}^{N_q\times d_v}$. Instead of relying on a learnable projection to determine the functionality of query tokens, we explicitly assign different roles to the queries. Specifically, the first $N_q/2=128$ query tokens $\mathbf{Q_0^s}\in\mathbb{R}^{{N_q}/2\times d_v}$  are designated as semantic queries, while the remaining $128$ queries $\mathbf{Q_0^a}\in\mathbb{R}^{{N_q}/2\times d_v}$ are designated as artifact queries. 
The query tokens are then concatenated with the visual tokens and fed into ViT, where they progressively aggregate visual information through stacked Transformer blocks. Consequently, the output embeddings are denoted as
$[\mathbf{V},\mathbf{Q}^{s},\mathbf{Q}^{a}]\in\mathbb{R}^{{(N_p+N_q)}\times d_v}$, where $N_p$ denotes the number of spatial patches.
\subsubsection{Decouple Head.}

We introduce a decouple head to transform the semantic and artifact query embeddings into task-specific latent representations. 
The semantic and artifact query embeddings are first projected into a shared latent space through  multi-layer perceptrons (MLPs), yielding
$\hat{\mathbf{Q}}^{s}\in\mathbb{R}^{N_q/2\times d_l}$ and
$\hat{\mathbf{Q}}^{a}\in\mathbb{R}^{N_q/2\times d_l}$,
where $d_l$ denotes the latent embedding dimension. The visual tokens are projected into the same latent space as
$\hat{\mathbf{V}}\in\mathbb{R}^{L\times d_l}$.
The query-conditioned features are generated using the inner product of query tokens $[\hat{\mathbf{Q}}^s,\hat{\mathbf{Q}}^a]$ and visual tokens $\hat{\mathbf{V}}$:
\begin{equation}
\boldsymbol{f}^{s}=\hat{\mathbf{V}}(\hat{\mathbf{Q}}^{s})^{\top},
\qquad
\boldsymbol{f}^{a}=\hat{\mathbf{V}}(\hat{\mathbf{Q}}^{a})^{\top},
\end{equation}
where
$\boldsymbol{f}^{s},\boldsymbol{f}^{a}\in\mathbb{R}^{N_p\times N_q/2}$.
The resulting features retain the spatial correspondence of visual tokens and integrate query-specific information. 
$\boldsymbol{f}^{s}$ captures semantic cues, while ${f}^{a}$ captures forgery-related artifact cues, which are further utilized for cross-reconstruction. 
For a pair of semantically aligned samples $(x_i,x_j)$, the decouple head produces four representations:
$\{\boldsymbol{f}^{s}_i,\boldsymbol{f}^{a}_i,\boldsymbol{f}^{s}_j,\boldsymbol{f}^{a}_j\}$.

To enhance the discrimination of artifact representations, we introduce a mask-guided patch-level contrastive objective.
Given a batch of artifact representations 
$\mathbf{F}^{a}\in\mathbb{R}^{B\times N_p\times d_l}$,
we further divide features into manipulated and non-manipulated regions according to the manipulation masks, denoted as $\mathbf{F}_{f}^{a}$ and $\mathbf{F}_{r}^{a}$, respectively.
Different from conventional contrastive learning that pulls all samples from the same category together, we do not impose alignment on manipulated patches. This is because artifacts introduced by different manipulation methods exhibit inherent diversity, and forcing them to be clustered may degrade the generalization ability to unseen generators.

Therefore, we encourage the real patches to maintain feature consistency, while separating forgery patches from real regions. The artifact contrastive loss is formulated as:
\begin{equation}
\mathcal{L}_{contra}
=
-\log
\frac{
\exp(\delta(\mathbf{F}_{r}^{a},\mathbf{F}_{r}^{a})/\tau)
}
{
\exp(\delta(\mathbf{F}_{r}^{a},\mathbf{F}_{r}^{a})/\tau)
+
\exp(\delta(\mathbf{F}_{f}^{a},\mathbf{F}_{r}^{a})/\tau)
}.
\end{equation}

\subsubsection{Decoder.}

To further enforce the disentanglement between semantic and artifact representations, we introduce a reconstruction decoder.
Given the four features produced by decouple head:
$\{\boldsymbol{f}^{s}_{i},\boldsymbol{f}^{a}_{i},\boldsymbol{f}^{s}_{j},\boldsymbol{f}^{a}_{j}\}$,
the decoder performs cross-reconstruction by exchanging artifact representations while preserving semantic representations. 
Specifically,  Adaptive Instance Normalization (AdaIN) ~\cite{huang2017arbitrary} is employed to inject artifact representations into semantic representations:
\begin{equation}
\mathrm{AdaIN}(\boldsymbol{f}^{s},\boldsymbol{f}^{a})
=
\sigma(\boldsymbol{f}^{a})
\frac{\boldsymbol{f}^{s}-\mu(\boldsymbol{f}^{s})}
{\sigma(\boldsymbol{f}^{s})}
+
\mu(\boldsymbol{f}^{a}),
\end{equation}
where $\mu(\cdot)$ and $\sigma(\cdot)$ denote the channel-wise mean and standard deviation, respectively. The exchanged representations are formulated as:
\begin{equation}
\mathbf{Z}_{j\leftarrow i}
=
\mathrm{AdaIN}(\boldsymbol{f}^{s}_{j},\boldsymbol{f}^{a}_{i}),
\qquad
\mathbf{Z}_{i\leftarrow j}
=
\mathrm{AdaIN}(\boldsymbol{f}^{s}_{i},\boldsymbol{f}^{a}_{j})
.
\end{equation}
The reconstructed images are obtained through a lightweight decoder as:
\begin{equation}
\hat{x}_{j\leftarrow i}=D(\mathbf{Z}_{j\leftarrow i}),
\qquad
\hat{x}_{i\leftarrow j}=D(\mathbf{Z}_{i\leftarrow j}),
\end{equation}
where $D$ denotes convolution layers that transform the dimension of the output as $\hat{x}\in\mathbb{R}^{h\times w\times 3}$ . 
\subsubsection{Reconstruction Target.}

Directly reconstructing RGB images may introduce excessive low-frequency semantic information into the reconstruction objective, making it difficult to explicitly optimize the manipulation-related artifacts. To alleviate this issue, we redefine the reconstruction target by removing the low-frequency components from the original image. 
A Gaussian filter is adopted as a low-frequency filter, and the reconstruction target is formulated as:
\begin{equation}
x^{gt}=x-\mathrm{G}(x),
\end{equation}
where $\mathrm{G}(\cdot)$ denotes the Gaussian blur operation. By suppressing low-frequency information, the reconstructed target mainly preserves high-frequency variations, where manipulation artifacts are more prominent.

Given the reconstructed target $x^{gt}$ and the prediction $\hat{x}$, the reconstruction loss is defined as:
\begin{equation}
\mathcal{L}_{rec}
=
\left\|
\hat{x}-x^{gt}
\right\| .
\end{equation}

Although the high-frequency reconstruction target emphasizes manipulation-related artifacts, not all high-frequency components are associated with forgery patterns. The background may contain irrelevant details, such as natural textures and scene details. Therefore, we introduce a spatial mask to focus the reconstruction supervision on manipulation-related regions. Since cross-reconstruction exchanges artifact representations between paired samples, the reconstructed target should contain artifact responses from both samples. We adopt the union of manipulation regions from the paired samples to cover all potential artifact areas. For a paired sample $(x_i,x_j)$, the reconstruction mask is defined as:
\begin{equation}
\mathcal{M}=\mathcal{M}_{i}\cup\mathcal{M}_{j},
\end{equation}
where $\mathcal{M}_{i}$ and $\mathcal{M}_{j}$ denote the binary pixel-level masks of the manipulated regions in the paired samples.

The final cross-reconstruction loss is calculated within the manipulated regions:
\begin{equation}
\mathcal{L}_{cross}
=
\mathcal{M}\odot(
\left\|
(\hat{x}_{j\leftarrow i}-x_i^{gt})
\right\| +
\left\|
(\hat{x}_{i\leftarrow j}-x_j^{gt})
\right\|)
,
\end{equation}
where $\odot$ denotes element-wise multiplication.
\subsubsection{Classification.}
To fully leverage the capability of the pretrained VLM, we adopt a frozen CLIP visual encoder with an attention bias mechanism similar to SideAdapter\cite{xu2023side} as our classification module. Specifically, the initial CLS token of CLIP is duplicated $N_q$ times to form additional tokens:
$\mathbf{X}_{[\mathrm{CLS}^{*}]}
=
[
\mathbf{X}_{[\mathrm{CLS}]},
\ldots,
\mathbf{X}_{[\mathrm{CLS}]}
]
\in
\mathbb{R}^{N_q\times d_c},$
which are concatenated with the original CLIP input tokens:
$[
\mathbf{X}^{0}_{\mathrm{vis}},
\mathbf{X}^{0}_{[\mathrm{CLS}]},
\mathbf{X}^{0}_{[\mathrm{CLS}^{*}]}].$
During the training process, the CLIP parameters are frozen, while the duplicated CLS tokens are updated through self-attention with an artifact-aware attention bias. Specifically, an attention mask is applied to the self-attention operation, where the duplicated CLS tokens $\mathbf{X}_{[\mathrm{CLS}^{*}]}$ are allowed to interact only with the visual patch tokens. At the $l$-th Transformer layer, the update of $\mathbf{X}_{[\mathrm{CLS}^{*}]}$ can be formulated as:
\begin{equation}
\mathbf{X}^{l+1}_{[\mathrm{CLS}^{*}]}
=
\mathrm{Softmax}
\left(
{
\mathbf{Q}^{l}_{[\mathrm{CLS}^{*}]}
(\mathbf{K}^{l}_{\mathrm{vis}})^{\top}
}
+
\mathbf{\Delta}^{\top}
\right)
\mathbf{V}^{l}_{\mathrm{vis}},
\end{equation}
where $\mathbf{Q}^{l}_{[\mathrm{CLS}^{*}]}$, $\mathbf{K}^{l}_{\mathrm{vis}}$, and $\mathbf{V}^{l}_{\mathrm{vis}}$ denote the query of $\mathbf{X}_{[\mathrm{CLS}^{*}]}$, key of visual tokens, and value of visual tokens at the $l$-th layer, respectively. $\mathbf{\Delta}$ represents the artifact-aware attention bias, which is generated from the decouple head using the same artifact queries $\mathbf{Q}^{a}$ as those used to generate the artifact representation. This process can be denoted as:
\begin{equation}
\mathbf{\Delta}
=
\phi(\mathbf{V})
(\hat{\mathbf{Q}}^{a})^{\top},
\end{equation}
where $\phi(\cdot)$ denotes a learnable projection implemented by MLPs.
After $L$ Transformer layers, the final duplicated CLS representations are obtained as $\mathbf{X}^{L}_{[\mathrm{CLS}^{*}]}
\in
\mathbb{R}^{N_q\times d_c}$.
The authenticity prediction is produced by applying a classification head:
\begin{equation}
\mathrm{\hat{y}}
=
\mathrm{Softmax}
(
\mathrm{Conv}
(
\mathbf{X}^{L}_{[\mathrm{CLS}^{*}]}
)
),
\end{equation}
where $\mathbf{\hat{y}}$ denotes the predicted class. Given the ground truth label $y$, the binary classification loss is formulated as:
\begin{equation}
    \mathcal{L}_{cls}=-\left(y\log\hat{y}+(1-y)\log(1-\hat{y})\right).
\end{equation}
\subsection{Overall Objectives}

The overall training objective is designed to jointly optimize the classification capability, artifact contrastive learning, and cross-reconstruction. 
The final objective is formulated as:
\begin{equation}
\mathcal{L}_{}
=
\lambda_1\mathcal{L}_{cls}
+
\lambda_2\mathcal{L}_{contra}
+
\lambda_3\mathcal{L}_{cross},
\end{equation}
where $\lambda_1$, $\lambda_2$, and $\lambda_3$ are hyper-parameters controlling the contribution of each loss term.
\section{Experiments}

\begin{table*}[t]
\centering
\begin{tabular}{c|c|ccccc|c}
\toprule
Method & Venue & CDF-v1 & CDF-v2 & DFDC & DFDCP & DFD & Avg.\\
\midrule
Xception~\cite{rossler2019faceforensics++}
& ICCV'19
&  0.779 & 0.737 & 0.708 & 0.737 & 0.816 & 0.755\\

F3Net~\cite{qian2020thinking}
& AAAI'20
& 0.777 & 0.735 & 0.702 & 0.735 & 0.798 & 0.749\\

X-ray~\cite{li2019face}
& CVPR'20
& 0.709 & 0.679 & 0.633 & 0.694 & 0.766 & 0.696\\

FFD~\cite{dang2020detection}
& CVPR'20
& 0.784 & 0.744 & 0.703 & 0.743 & 0.802 & 0.755\\

SPSL~\cite{liu2021spatial}
& CVPR'21
& 0.815 & 0.765 & 0.704 & 0.741 & 0.812 & 0.767\\

SRM~\cite{luo2021generalizing}
& CVPR'21
& 0.793 & 0.755 & 0.700 & 0.741 & 0.812 & 0.760\\

Recce~\cite{cao2022end}
& CVPR'22
& 0.768 & 0.732 & 0.713 & 0.734 & 0.812 & 0.752\\

SBI~\cite{shiohara2022detecting}
& CVPR'22
& - & 0.813 & - & 0.799 & 0.774 & -\\

UCF~\cite{yan2023ucf}
& ICCV'23
& 0.779 & 0.753 & 0.719 & 0.759 & 0.807 & 0.763\\

ED~\cite{ba2024exposing}
& AAAI'24
& 0.818 & 0.864 & 0.721 & 0.851 & - & -\\

LSDA~\cite{yan2024transcending}
& CVPR'24
& 0.867 & 0.830 & 0.736 & 0.815 & 0.880 & 0.826\\

CFM~\cite{luo2023beyond}
& TIFS'24
& - & 0.828 & - & 0.758 & 0.915 & -\\

UDD~\cite{fu2025exploring}
& AAAI'25
& - &0.869 & 0.758&  0.856& 0.910& -\\

ForAda~\cite{cui2025forensics}$^{\dagger}$
& CVPR'25
& \underline{0.909} &0.882 &  0.861& \underline{0.905}& \underline{0.924} & 0.896\\

CRDA~\cite{chou2026improving}
& AAAI'26
&  0.901 &0.854  & 0.743& 0.797&  -& -\\

VLAForge~\cite{zhu2026unleashing}$^{\dagger}$
& CVPR'26
& 0.907 &\underline{0.895}  & \underline{0.869}& \textbf{0.907} &\textbf{0.929} & \underline{0.901}\\

\midrule

\textbf{Ours}
& -
& \textbf{0.930}
& \textbf{0.903}
& \textbf{0.873}
& {0.898}
& \textbf{0.929}
& \textbf{0.906}
\\

\bottomrule
\end{tabular}
\caption{Cross-dataset evaluation results (Frame-level AUC). 
All methods are trained on FF++ and evaluated on other datasets. 
The best results are indicated in bold and the \underline{second-best} results are underlined.
$^{\dagger}$ denotes results reproduced by us.}
\label{tab:cross_dataset}

\end{table*}

\begin{table}[t]
\centering
\small

\setlength{\tabcolsep}{4pt}
\begin{tabular}{c|c|cc}
\toprule
Method & Venue & CDF-v2 & DFDC\\
\midrule


TALL~\cite{xu2023tall}
& ICCV'23
& 0.908 & 0.768 \\


SeeABLE~\cite{larue2023seeable}
& ICCV'23
& 0.873 & 0.759 \\


TALL++~\cite{xu2024learning}
& IJCV'24
& 0.920 & 0.785 \\

SAM~\cite{choi2024exploiting}
& CVPR'24
& 0.890 & - \\

FCG~\cite{han2025towards}
& CVPR'25
& \underline{0.950} & 0.818 \\

\midrule


AUNet~\cite{bai2023aunet}
& CVPR'23
& 0.928 & 0.738 \\

CADDM~\cite{dong2023implicit}
& CVPR'23
& 0.939 & 0.739 \\

SFDG~\cite{wang2023dynamic}
& CVPR'23
& 0.758 & 0.736 \\

LSDA~\cite{yan2024transcending}
& CVPR'24
& 0.911 & 0.770 \\


ForAda~\cite{cui2025forensics}$^{\dagger}$
& CVPR'25
& 0.929 & 0.883 \\

RepDFD~\cite{lin2025standing}
& AAAI'25
& 0.899 & 0.810 \\

VLAForge~\cite{zhu2026unleashing}$^{\dagger}$
& CVPR'26
& 0.942 & \underline{0.893} \\

\midrule

\textbf{Ours}
& -
& \textbf{0.951}
& \textbf{0.895}
\\

\bottomrule
\end{tabular}
\caption{Cross-dataset evaluation results (Video-level AUC). 
The top section presents video-based methods and the bottom section corresponds to frame-based methods.
$^{\dagger}$ denotes results reproduced by us.}
\label{tab:video_auc}
\end{table}

\begin{table}[t]
\centering
\resizebox{\linewidth}{!}{
\begin{tabular}{c|c|cccc}
\toprule
\textbf{Setting} & \textbf{Method} & \textbf{VQGAN} & \textbf{StyleGAN} & \textbf{SiT} & \textbf{DiT} \\
\midrule
\multirow{3}{*}{F-level}

& ForAda$^{\dagger}$  & 0.978 & 0.897 & 0.767 & 0.707 \\
& VLAForge$^{\dagger}$ & 0.983 & \textbf{0.980} & 0.794 & 0.737 \\
& \textbf{Ours} & \textbf{0.984} & 0.958 & \textbf{0.802} & \textbf{0.749} \\
\midrule
\multirow{3}{*}{V-level}

& ForAda$^{\dagger}$ & 0.998 & 0.965 & 0.850 & 0.800 \\
& VLAForge$^{\dagger}$& 0.995 & 0.993 & 0.880 & 0.815 \\
& \textbf{Ours} & \textbf{0.999} & \textbf{0.995} & \textbf{0.881} & \textbf{0.824} \\
\bottomrule
\end{tabular}
}
\caption{AUC results on frame-level (F-level) and video-level (V-level) forgery detection across GAN- and diffusion-based generators. $^{\dagger}$ denotes results reproduced by us.}
\label{tab:generator_detection}
\end{table}

\subsection{Settings}
\subsubsection{Dataset.}
Following previous works, we train our model on the c23 version of FaceForensics++ (FF++) dataset\cite{rossler2019faceforensics++}. 
For evaluation, we conduct cross-dataset experiments on widely-used benchmarks, including Celeb-DF-v1/v2 \cite{li2020celeb}, DFDCP \cite{dolhansky2019deepfake}, DFDC \cite{dolhansky2020deepfake}, DFD \cite{dfd}, and four full-face synthesis datasets in DF40 \cite{yan2024df40}, including VQGAN \cite{esser2021taming}, StyleGAN-XL (StyleGAN) \cite{sauer2022stylegan}, SiT-XL/2 (SiT) \cite{atito2021sit}, and DiT \cite{peebles2023scalable}.
\subsubsection{Implementation Details.}
We use ViT-Tiny/16~\cite{dosovitskiy2020image} for visual feature extraction and a frozen CLIP ViT-L/14~\cite{radford2021learning} for classification. The entire model is trained on the FF++ c23 training set with a batch size of 16, using the Adam optimizer~\cite{kingma2014adam} with a learning rate of 0.0002 and a weight decay of 0.0005. The overall objective is optimized with a weighted combination of classification loss, artifact contrastive loss, and cross-reconstruction loss, where the loss weights $\lambda_1$, $\lambda_2$, and $\lambda_3$ are set to 10, 20, and 50, respectively. Detection performance is evaluated by AUC (Area Under the ROC Curve), AP (Average Precision), and EER (Equal Error Rate).The random seed is fixed to 1020 for reproducible experiments. All experiments are conducted on one NVIDIA RTX 4090 GPU.
\subsection{Evaluations}
\subsubsection{Cross-Dataset Evaluation.}

\begin{table*}[t]
\centering
\small
\setlength{\tabcolsep}{3pt}

\begin{tabular}{cc|ccc|ccc|ccc|ccc|ccc}
\toprule

\multirow{2}{*}{$\mathcal{L}_{contra}$}
&
\multirow{2}{*}{$\mathcal{L}_{cross}$}
&
\multicolumn{3}{c|}{CDF-v1}
&
\multicolumn{3}{c|}{CDF-v2}
&
\multicolumn{3}{c|}{DFDC}
&
\multicolumn{3}{c|}{DFDCP}
&
\multicolumn{3}{c}{DFD}
\\

\cmidrule(lr){3-5}
\cmidrule(lr){6-8}
\cmidrule(lr){9-11}
\cmidrule(lr){12-14}
\cmidrule(lr){15-17}

&
&
AUC $\uparrow$ & AP $\uparrow$ & EER $\downarrow$
&
AUC $\uparrow$ & AP $\uparrow$ & EER $\downarrow$
&
AUC $\uparrow$ & AP $\uparrow$ & EER $\downarrow$
&
AUC $\uparrow$ & AP $\uparrow$ & EER $\downarrow$
&
AUC $\uparrow$ & AP $\uparrow$ & EER $\downarrow$
\\

\midrule

$\times$ & $\times$
& 0.856 & 0.914 & 23.2
& 0.853 & 0.920 & 23.3
& 0.860 & 0.886 & 22.0
& 0.894 & 0.943 & 19.2
& 0.920 & 0.982 & 14.6
\\

$\times$ & $\checkmark$
& 0.914 & 0.939 & 18.4
& 0.893 & 0.940 & \textbf{13.1}
& 0.858 & 0.883 & 22.5
& 0.891 & 0.941 & 19.5 
& 0.922 & 0.989 & 13.9
\\

$\checkmark$ & $\times$
& 0.889 & 0.928 & 19.5
& 0.869 & 0.927 & 21.8
& 0.846 & 0.874 & 23.5
& 0.876 & 0.935 & 20.8
& 0.918 & 0.990 & 14.7 
\\

$\checkmark$ & $\checkmark$
&\textbf{0.930}&\textbf{0.955}&\textbf{15.6}
&\textbf{0.903}&\textbf{0.947}&18.4
&\textbf{0.873}&\textbf{0.897}&\textbf{21.1}
&\textbf{0.898}&\textbf{0.945}&\textbf{18.6}
&\textbf{0.929}&\textbf{0.991}&\textbf{13.8}
\\

\bottomrule
\end{tabular}

\caption{
Ablation study on the effect of contrastive and cross-reconstruction losses.
Evaluation metrics are frame-level AUC, AP, and EER, respectively.
}
\label{tab:loss_ablation}

\end{table*}

\begin{table*}[t]
\centering
\small
\setlength{\tabcolsep}{3pt}

\begin{tabular}{ccc|ccc|ccc|ccc|ccc|ccc}
\toprule

\multirow{2}{*}{Setting}
&
\multirow{2}{*}{Pair}
&
\multirow{2}{*}{Target}
&
\multicolumn{3}{c|}{CDF-v1}
&
\multicolumn{3}{c|}{CDF-v2}
&
\multicolumn{3}{c|}{DFDC}
&
\multicolumn{3}{c|}{DFDCP}
&
\multicolumn{3}{c}{DFD}
\\

\cmidrule(lr){4-6}
\cmidrule(lr){7-9}
\cmidrule(lr){10-12}
\cmidrule(lr){13-15}
\cmidrule(lr){16-18}

&
&
&
AUC $\uparrow$ & AP $\uparrow$ & EER $\downarrow$
&
AUC $\uparrow$ & AP $\uparrow$ & EER $\downarrow$
&
AUC $\uparrow$ & AP $\uparrow$ & EER $\downarrow$
&
AUC $\uparrow$ & AP $\uparrow$ & EER $\downarrow$
&
AUC $\uparrow$ & AP $\uparrow$ & EER $\downarrow$
\\

\midrule

\textbf{Conv.}
& R-F
& RGB
&0.871&0.918&23.2
&0.841&0.907&23.6
&0.818&0.835&26.2
&0.845&0.911&24.2
&0.903&0.988&17.9
\\

& F-F
& RGB
&0.905&0.941&18.0
&0.896&0.942&18.9
&0.869&0.893&\textbf{21.1}
&\textbf{0.908}&0.943&\textbf{17.1}
&0.925&0.989&14.1
\\

&
R-F
& Res
&
0.886&0.932&20.9
&0.880&0.936&20.8
&0.859&0.883&21.9
&0.869&0.930&21.9
&0.924&0.991&13.9
\\

\textbf{Ours}
& F-F
& Res
&
\textbf{0.930}&\textbf{0.955}&\textbf{15.6}
&
\textbf{0.903}&\textbf{0.947}&18.4
&
\textbf{0.873}&\textbf{0.897}&\textbf{21.1}
&
0.898&\textbf{0.945}&18.6
&
\textbf{0.929}&\textbf{0.991}&\textbf{13.8}
\\

\bottomrule
\end{tabular}

\caption{Ablation study on the cross-reconstruction module. 
R-F and F-F denote real-fake pairs and semantically identical fake pairs. 
RGB and Res represent RGB image reconstruction target and our residual target.}
\label{tab:task_ablation}

\end{table*}

We first evaluate the generalization ability of our method under a cross-dataset setting, where models are trained on FF++ and tested on several unseen datasets. As shown in Table~\ref{tab:cross_dataset}, our method achieves the best overall performance among compared approaches, obtaining the highest average AUC of 90.6\% across five benchmark datasets. In particular, our method improves the detection performance on multiple unseen domains, achieving AUC scores of 93.0\% and 90.3\% on CDF-v1 and CDF-v2, respectively, while also maintaining strong performance on DFDC, DFDCP, and DFD. These results demonstrate that our proposed representation learning strategy can effectively capture generalizable forensic patterns beyond dataset-specific artifacts, leading to more robust cross-domain deepfake detection.

We further evaluate whether the learned representations can be effectively generalized from frame-level detection to video-level prediction. 
Comparison with existing approaches is presented in Table~\ref{tab:video_auc}. 
Notably, our method achieves state-of-the-art performance on CDF-v2 and DFDC, obtaining AUC scores of 95.1\% and 89.5\%, respectively. 

These results further verify that the learned representations can capture robust forensic cues beyond specific datasets and remain effective when applied to practical video-level deepfake detection scenarios.

\subsubsection{Cross-Generator Evaluation.}
To evaluate generalization to unseen generation methods, we conduct cross-dataset experiments on four unseen generators, including VQGAN, StyleGAN, SiT, and DiT. As shown in Table~\ref{tab:generator_detection}, our method achieves competitive performance at both frame and video levels. Particularly, our method outperforms existing approaches on diffusion-based generators, achieving AUC improvements on SiT and DiT. This demonstrates that preserving artifact diversity enables the model to capture complementary forensic cues beyond seen generator statistics, improving generalization to unseen manipulations. For GAN-based generators, our method also achieves comparable performance.

\subsection{Ablation Study}
\subsubsection{Impact of Loss Functions.}
We investigate the effectiveness of the proposed loss functions in Table~\ref{tab:loss_ablation}. 
Since the classification loss directly supervises the final deepfake detection objective, we keep it unchanged and focus on evaluating the contributions of $\mathcal{L}_{contra}$ and $\mathcal{L}_{cross}$. 
Removing both losses results in a clear performance drop, demonstrating the importance of learning artifact-aware representations. 
Among them, $\mathcal{L}_{cross}$ provides the dominant improvement, increasing the AUC from 85.6\% to 91.4\% on CDF-v1 and from 85.3\% to 89.3\% on CDF-v2. 

\subsubsection{Impact of Cross-Reconstruction Settings.}

We investigate the impact of different cross-reconstruction designs, including the choice of reconstruction pairs and targets. 
As shown in Table~\ref{tab:task_ablation}, we establish a baseline using real-fake pairs with RGB image reconstruction. 
By replacing real-fake pairs with semantically identical fake pairs generated by different manipulation methods, the detection performance is improved on most datasets. 
The AUC on CDF-v1 increases from 87.1\% to 90.5\%, demonstrating that fake-fake pairs provide a more effective way to capture diverse forensic artifacts while reducing semantic interference.

Furthermore, replacing RGB reconstruction with residual reconstruction further enhances the generalization capability by focusing on high-frequency forensic cues rather than image content. 
Combining both designs, i.e., fake-fake pairs with residual reconstruction, achieves the best overall performance, obtaining the highest AUC on four out of five benchmark datasets and improving the AUC on CDF-v1 from 87.1\% to 93.0\%. 
These results verify that the proposed cross-reconstruction strategy benefits from both artifact-aware pair construction and residual-level supervision, enabling the model to learn more generalizable forensic representations beyond dataset-specific appearance information.

\subsection{Visualization}
\subsubsection{The t-SNE Visualization.}
\begin{figure}[t]
    \centering
    \includegraphics[width=\linewidth]{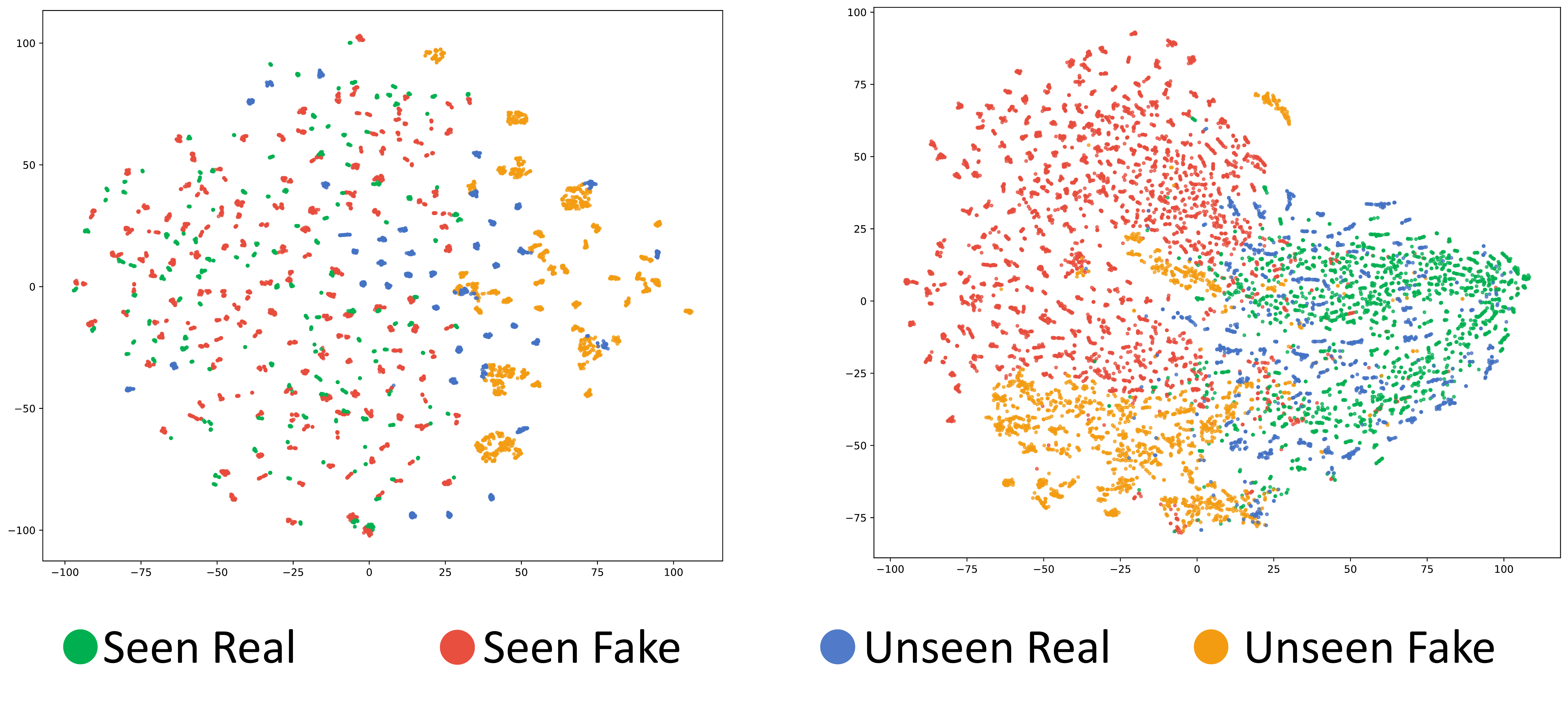}
    \caption{
    T-SNE visualization of feature distributions.
    }
    \label{fig:tsne}
\end{figure}

To further analyze the effectiveness of the proposed adapter, we visualize the feature distributions extracted from the original CLIP and the adapted CLIP with the proposed attention bias mechanism. Specifically, we construct a visualization set containing 2,000 samples, where the seen samples are randomly selected from the FF++, while the unseen samples are collected from the CDF-v1. We then apply the t-SNE algorithm to project the extracted features into a two-dimensional space for visualization.
As shown in Fig.~\ref{fig:tsne}, 
Our method does not require explicit feature alignment between different datasets, yet still preserves clear discrimination under domain shifts. This further supports our motivation that modeling diverse and transferable forensic artifacts is more effective than enforcing dataset-level alignment for cross-domain generalization.

\section{Conclusion}
In this work, we revisit cross-reconstruction for image forgery detection and highlight the importance of artifact diversity in improving generalization. 
Unlike previous approaches that attempt to align heterogeneous artifacts or exclude them from reconstruction, we demonstrate that diverse manipulation artifacts contain complementary forensic cues that should be preserved and exploited.
Based on this insight, we propose an artifact-aware disentanglement framework that leverages semantically aligned cross-generator pairs. 
By incorporating artifact representations into cross-reconstruction and introducing masked high-frequency residual targets, our framework learns transferable forensic representations while maintaining diverse artifact characteristics.
Extensive experiments on multiple benchmarks demonstrate the effectiveness of our approach under both cross-dataset and cross-generator evaluation settings.

\appendix

\bigskip

\newpage
\bibliography{aaai2027}


\end{document}